\documentclass[lettersize,journal]{IEEEtran}
\usepackage{amsmath,amsfonts}
\usepackage{algorithmic}
\usepackage{array}
\usepackage[caption=false,font=normalsize,labelfont=sf,textfont=sf]{subfig}
\usepackage{textcomp}
\usepackage{stfloats}
\usepackage{url}
\usepackage{verbatim}
\usepackage{graphicx}
\usepackage{algorithm}
\usepackage{algorithmic}
\usepackage{booktabs}
\usepackage{cite}

\def\BibTeX{{\rm B\kern-.05em{\sc i\kern-.025em b}\kern-.08em
    T\kern-.1667em\lower.7ex\hbox{E}\kern-.125emX}}
\usepackage{balance}
\begin{document}
\title{MapTrack: Tracking in the Map}
\author{Fei Wang, Ruohui Zhang, Chenglin Chen, Min Yang, Yun Bai
\thanks{Fei Wang, Ruohui Zhang, Chenglin Chen, and Min Yang are with the Thrust of Intelligent Transportation, The Hong Kong University of Science and Technology
(Guangzhou), Guangzhou, China (e-mail:\{fwang423, rzhang856, cchen363, myang945\}@connect.hkust-gz.edu.cn).

Yun Bai is with the Thrust of Intelligent Transportation and Guangdong
Provincial Key Lab of Integrated Communication, Sensing and Computation
for Ubiquitous Internet of Things, The Hong Kong University of Science
and Technology (Guangzhou), Guangzhou, China; the State Key Laboratory of Rail Traffic Control and Safety, Beijing Jiaotong University,
Beijing, China (e-mail: yunbai@hkust-gz.edu.cn).
}}

\markboth{}%
{Shell \MakeLowercase{\textit{et al.}}: A Sample Article Using IEEEtran.cls for IEEE Journals}

\maketitle

\begin{abstract}
  Multi-Object Tracking (MOT) aims to maintain stable and uninterrupted trajectories for each target. 
  Most state-of-the-art approaches first detect objects in each frame and then implement data association 
  between new detections and existing tracks using motion models and appearance similarities. 
  Despite achieving satisfactory results, occlusion and crowds can easily lead to missing and distorted 
  detections, followed by missing and false associations. In this paper, we first revisit the classic tracker DeepSORT, 
  enhancing its robustness over crowds and occlusion significantly by 
  placing greater trust in predictions when detections are unavailable or of low quality in 
  crowded and occluded scenes. Specifically, we propose a new framework comprising of three lightweight and plug-and-play 
  algorithms: the probability map, 
  the prediction map, and the covariance adaptive Kalman filter.
  The probability map identifies whether undetected objects have genuinely disappeared from 
  view (e.g., out of the image or entered a building) or are only temporarily undetected 
  due to occlusion or other reasons. 
  Trajectories of undetected targets that are still within
  the probability map are extended by state estimations directly. The prediction map determines whether an object is in a crowd, 
  and we prioritize state estimations over observations when severe deformation of observations 
  occurs, accomplished through the covariance adaptive Kalman filter. The proposed method, 
  named MapTrack, achieves state-of-the-art results on popular multi-object tracking 
  benchmarks such as MOT17 and MOT20. Despite its superior performance, our method 
  remains simple, online, and real-time. The code will be open-sourced later.
\end{abstract}


\section{Introduction}
\IEEEPARstart{M}{ulti-object tracking (MOT)} is dedicated to detecting and tracking specific
classes of objects 
frame by frame, a crucial task with applications in fields such as autonomous driving. 
In recent years, the tracking-by-detection (TBD) paradigm~\cite{sort, deepsort, zhang2022bytetrack, 
bochinski2017high, centertrack, tracktor} has dominated the MOT task due 
to the significant advancements in object detection~\cite{yolox, girshick2015fast, redmon2016you, redmon2017yolo9000}. This paradigm involves detecting objects 
in each frame and subsequently employing a matching algorithm, such as the Hungarian algorithm~\cite{munkres1957algorithms}, 
to associate these detections with previously confirmed tracks.

DeepSORT serves as a prototypical and robust exemplar within the realm of Tracking-by-Detection (TBD) 
methodologies, and a myriad of notable contributions~\cite{fairmot, zhang2022bytetrack, du2023strongsort, giaotracker} 
have emanated from the foundation laid by 
DeepSORT, achieving the state-of-the-art performance. DeepSORT-like Frameworks
incorporate an optimal filtering mechanism, exemplified by the Kalman filter~\cite{kalman1960new} and its 
various iterations, to manage tracking states that encapsulate historical information. The 
Kalman filter includes two stages: the prediction stage and the update stage. 
During the prediction stage, object states on the succeeding frame are predicted through 
a transition function, typically grounded in the assumption of constant velocity, 
yielding "predictions" alongside their corresponding uncertainties, represented 
by the prediction covariance. The subsequent update stage harnesses object state 
measurements from the object detector  
to refine the predictions using a weighted average, which is called the Kalman gain.
The greater the uncertainty in measurements, represented by the measurement covariance, 
the higher the weight assigned to predictions.
The associations between predictions and newly detected objects are established through matching 
algorithms utilizing distance metrics such as Intersection over Union (IoU) and the appearance similarity.

Nonetheless, a prevalent issue encountered in DeepSORT-like frameworks
is the phenomenon of ID switching, wherein the same objects are identified as different IDs 
in consecutive frames, which often happens in 
challenging scenes such as crowds, intersections of objects, and occlusion.
According to our extensive experimental results, the specific reasons of ID switching in 
these scenes for DeepSORT could be summarized as below:
\begin{itemize}
  \item [$\bullet$]In the context of crowded scenes where ground-truth bounding boxes exhibit overlap, 
  three scenarios commonly arise. Firstly, objects may be entirely or partially occluded, 
  resulting in their detections being overlooked. In such instances, DeepSORT-like methodologies
  usually expunge unmatched identities, either immediately or upon the tracker's determination 
  that the objects have exited the frame. Consequently, upon re-emergence, these objects are 
  assigned new identities. Secondly, the detected bounding boxes may be erroneously enlarged or partially detected due to overlap, thereby becoming more easily associated with
  incorrect tracks. Thirdly, enlarged or partially detected objects may be linked 
  to the correct tracks, compelling the trajectories to be 
  deviated seriously because the deformation of the detections leads to the saltation of the updated 
  states in the update stage of the Kalman filter.
  \item [$\bullet$]The association between detections and tracks only utilizes the information of two consecutive frames, not taking advantage of global information.
  \item [$\bullet$]Similar appearance confuses the re-identification (ReID) model.
  \item [$\bullet$]Matching cascade algorithm of DeepSORT gives greater matching priority to more frequently seen objects, which is unreasonable in many cases.
\end{itemize}

In summary, methodologies grounded in DeepSORT often fall short in adeptly addressing scenarios 
characterized by the crowd and occlusion, and the reliance on localized and frame-by-frame associations restricts the exploitation of global information. Consequently, a lot of works try to improve the performance of MOT 
by improving the discriminability of the ReID model~\cite{pang2021quasi, wang2021multiple, ristani2018features, fischer2023qdtrack, maggiolino2023deep}
or paying attention to the motion model
~\cite{observation_centric, giaotracker, du2023strongsort, yang2023hard} to discriminate between pedestrians, especially in highly crowded scenes.

With the rise and ascendancy of deep learning, numerous researchers endeavor to harness 
the capabilities of deep learning models for enhanced data association. Some works~\cite{carion2020end, meinhardt2022trackformer, sun2020transtrack, zhou2022global} 
do association based on the attention mechanism of the transformer. Some other
works~\cite{tang2016multi, tang2017multiple, tang2015subgraph, wen2014multiple, braso2020learning, he2021learnable, wang2019exploit} 
utilize Graph Convolutional Network (GCN) to do data association by considering detections 
or tracklets as nodes and the link between two nodes as an edge. 
Even though some of them achieve very competitive results, their practical utility remains 
circumscribed for several 
reasons. Firstly, most of them depend on heavy-weight deep learning architectures, which are 
too resource-hungry and computationally
expensive to achieve real-time performance, especially on 
low computing power platforms. Secondly, the inherent 
data-driven nature of deep learning models necessitates retraining for optimal performance 
across different scenes. Thirdly, the deep learning models are fixed and work like black boxes somewhat, 
resulting in the lack of scalability to the actual scenes.

Thus, we revisit DeepSORT for its deployability, simplicity, scalability, and effectiveness. 
It is noteworthy to acknowledge that the underperformance of DeepSORT, when compared with contemporary 
state-of-the-art methodologies, stems from the obsolescence of its techniques rather than any 
inherent limitations in its tracking paradigm~\cite{du2023strongsort}. Meanwhile, we observe a resurgence 
of interest in the 
motion model~\cite{sun2022dancetrack, cioppa2022soccernet, qin2023motiontrack, du2023strongsort, observation_centric, yang2023hard}. 
Some methods even abandoned appearance information completely~\cite{han2022mat, hu2023stdformer, fischer2023qdtrack}, relying only on 
motion information and achieving high running speed and competitive performance. 
However, appearance similarity is a useful information in most scenes, and abandoning
appearance features would decrease the robustness in complex scenes.
Therefore, in this paper, we adopt the DeepSORT-like paradigm and place more emphasis on 
the handling of occlusion and crowds to improve the robustness of this classic framework.

The motivations and rationales of our methods can be summarized as follows:
\begin{itemize}
  \item [$\bullet$]We should trust detections more than predictions when objects are not occluded or in the crowd owing to the following three inherent 
  limitations associated with the constant velocity assumption~\cite{observation_centric}.
  Firstly, it doesn't work well when confronted with non-linear motion patterns.
  Secondly, it is sensitive to the noise of the displacement of the object between adjacent frames due to the small time interval,
  leading to the saltation of the estimated velocity by the Kalman filter.
  Thirdly, the error in the velocity estimate translates into the error in the position estimate, accumulating rapidly in the absence of consecutive detections.
  \item [$\bullet$]Numerous factors contribute to the non-detection of a normal target in a given frame during subsequent frames.
  For example, the target vanishes from the visual field really, such as exiting the image or entering a building.
  Or it may be occluded by other things temporarily. Also it may be due to a false negative of the detector. 
  However, we need to know whether it really vanishes from the view or not,
  which is judged by the probability map.
  If a track is in the probability map and there is no detection matched for it, 
  it is highly likely that it is occluded or the detector fails to detect it, so the predicted position
  of the object in the incoming
  frame is employed directly. We also develop corresponding filtering algorithms to delete the wrongly predicted tracks.
  \item [$\bullet$]In cases where a track is situated within a crowd, which is evaluated by the prediction 
  map, and substantial deformation is observed in the associated detected bounding box, 
  we should not trust the detection as much 
  as in the uncrowded scenes. To address this challenge, a Covariance Adaptive Kalman 
  filter (CA Kalman filter) is devised, which is capable of adaptively 
  adjusting measurement covariance based on the quality of object detections.
  \item [$\bullet$]Considering the sensitivity to the noise inherent in the constant velocity assumption as mentioned 
  above, a momentum strategy is employed to smooth out the velocity measurements. 
  This approach holds substantial significance for enhancing the robustness of our prediction-based strategy.
  \item [$\bullet$]It is a common occurrence for an object to exit the frame or become occluded and then reappear. 
   We tackle this problem by memorizing the tracklets not in the probability map and re-identifying them by the ReID model upon their reappearance.
\end{itemize}

To let the object to continue in the direction it followed before,
we actually utilize the historical trajectory information, which could also be viewed as spatiotemporal constraints.
Our light-weight and effective framework, named MapTrack for tracking in the probability map and prediction map, 
remains simple, online, real-time and significantly improves robustness over occlusion and crowds.
The principal contributions of our work can be succinctly highlighted as follows:
\begin{itemize}
  \item [$\bullet$]We propose an innovative tracking framework, named MapTrack, which excels at addressing the challenges associated with 
  occlusion and crowds. It achieves state-of-the-art performance on multiple
  datasets in an online and real-time fashion. It can serve as a stronger baseline for tracking-by-detection trackers
  and motivate more concern on non-deep learning association methods and motion models.
  \item [$\bullet$]We develop the probability map, the prediction map, and the CA Kalman filter to 
  address challenges related to occlusion and crowded scenarios specifically. Furthermore, they also can be integrated into other trackers 
  easily and benefit some tracking-based tasks in both academic and industrial contexts.
\end{itemize}

\section{Related Works}
\subsection{DeepSORT-like Paradigm}
\noindent As the precursor of DeepSORT, SORT~\cite{sort} applies a Kalman filter for each tracked instance, predicting its state in the subsequent frame and utilizing the 
IoU as the distance metric to associate the predicted bounding boxes to the detected boxes. 
Subsequently, as deep networks revolutionize the extraction of high-quality visual features,
DeepSORT~\cite{deepsort} seamlessly integrates appearance information with SORT by employing a pre-trained ReID model. This pragmatic and effective tracking paradigm has been adopted by
many following works~\cite{zhang2022bytetrack, yu2016poi, du2023strongsort}. 
ByteTrack~\cite{zhang2022bytetrack} harnesses the capabilities of a powerful YOLOX-based~\cite{yolox} detector and introduces an improved SORT algorithm 
to facilitate the association of not only high-scoring detection boxes but also those with lower scores, resulting in state-of-the-art performance.
DeepSORT is improved by StrongSORT~\cite{du2023strongsort} through the integration of 
two simple algorithms: AFLink and GSI. 
GSI, a lightweight interpolation algorithm, utilizes
Gaussian process regression~\cite{williams1995gaussian} to model nonlinear motion, filling the gaps in trajectories stemming from missing detections. 
AFLink, a Convolutional Neural Network (CNN) model, is designed to further filter out unreasonable tracklet pairs by the spatiotemporal features.
Given the time-consuming nature of both the individual detector and ReID models, and considering 
the potential for the ReID model to share the same low-level model structure as the detector, 
some works~\cite{jde, fairmot} consolidate these two components into a unified one.
JDE~\cite{jde} introduces a unified detection and ReID model that concurrently generates detection 
results along with the corresponding appearance embeddings for the detected boxes.
FairMOT~\cite{fairmot} identifies three limitations in previous combinations of detection and ReID, proposing an enhanced one-shot tracker based on CenterNet~\cite{zhou2019objects}.
The primary benefits of these integrated trackers are low computational costs and not bad performance.

\subsection{Deep Leaning Based Association}
\noindent With recent advancements in deep learning and the increased accessibility of computing power, many studies employ deep learning models to address challenges associated with missing and false associations.
Numerous studies conceptualize data association as a graph-based optimization problem and solve it by minimizing the overall 
cost. In~\cite{tang2016multi, tang2017multiple} the vertices are the detections, while in~\cite{tang2015subgraph, wen2014multiple} 
the vertices are tracklets.
Certain methodologies integrate detection and data association into a unified model, transforming the detector 
into a  tracker~\cite{tracktor, centertrack, voigtlaender2019mots}. 
Tractor~\cite{tracktor} utilizes the regression head of the Faster R-CNN~\cite{fasterrcnn} to produce the predicted bounding box of
the object in the subsequent frame.
CenterTrack~\cite{centertrack} incorporates previous frame and a heatmap of prior tracklets into CenterNet~\cite{zhou2019objects}
as supplementary input channels. This enables the model 
to predict inter-frame offsets between the centers of the objects.
Inspired by the successful applications of transformer architecture~\cite{vaswani2017attention} in vision-related tasks, 
certain works~\cite{carion2020end, meinhardt2022trackformer, sun2020transtrack, zhou2022global} have adapted it to MOT tasks,
resulting in good performance.
DETR~\cite{carion2020end} is an end-to-end trainable MOT framework based on the Transformer architectures. It utilizes a CNN backbone to extract features
and a simple feed forward network (FFN) to predict the bounding boxes which are associated with detections by a bipartite matching algorithm.
TrackFormer~\cite{meinhardt2022trackformer} enhances DETR by incorporating existing tracks as input for the Transformer,
facilitating the joint execution of detection, tracking, and data association.
TransTrack~\cite{sun2020transtrack} leverages an attention-based query-key mechanism to associate 
object queries with track queries through a straightforward IoU matching process.

\subsection{Occlusion Handling}
\noindent How to deal with occlusions has been a persistent challenge in MOT. 
A common method to mitigate occlusion is to use appearance information to discriminate objects~\cite{deepsort, pang2021quasi, wang2021multiple, ristani2018features, fischer2023qdtrack, maggiolino2023deep}. 
Nonetheless, the discriminability of ReID models relies excessively on detector performance, 
and it typically decreases significantly in scenes characterized by blurriness, partial detection, crowds, and similar appearances.  

To tackle this challenge more effectively, various methods~\cite{giaotracker, qin2023motiontrack, wang2019exploit, tpm, remot, tang2015subgraph, wen2014multiple, zhou2022global}
exploit rich global information. Typically, accurate yet incomplete tracklets are initially generated using spatiotemporal and/or appearance information,
followed by their association with trajectories. 
A multi-scale TrackletNet is proposed by
TNT~\cite{wang2019exploit} to assess the similarity between tracklets.
This enables the reconnection of broken tracklets caused by unreliable detections 
and occlusions using a graph model, with tracklets serving as vertices.
GIAOTracker~\cite{giaotracker} introduces a global link algorithm that utilizes the GIModel,
which is based on ResNet50-TP~\cite{gao2018revisiting}, 
to extract appearance features from tracklets.
It establishes connections between tracklets by minimizing both appearance and spatiotemporal distances. 
While these global link methods exhibit significant improvements, they often depend 
on computationally expensive deep learning models and involve numerous 
hyperparameters for fine-tuning. Additionally, an excessive 
reliance on appearance features can render the approach susceptible to occlusion.

An alternative strategy to compensate for occlusion is to develop more robust motion models 
particularly tailored for occlusion scenarios~\cite{observation_centric, qin2023motiontrack, han2022mat, yang2021learning}. 
MotionTrack~\cite{qin2023motiontrack} incorporates an Interaction Module to acquire interaction-aware motions 
from short-term trajectories and a refined module to capture reliable long-term motions from the target's new
history trajectory. This design enables the handling of short-term and long-term occlusions, respectively.
MAT~\cite{han2022mat} designs the motion-based dynamic reconnection context (DRC) module to reconnect the deactivated trajectories,
and the 3D Integral Image (3DII) to eliminate erroneous connections  
through spatiotemporal constraints during the data association stage. 

\section{Review of DeepSORT}
\noindent DeepSORT manages tracks and estimates states using a Kalman filter based on the linear constant velocity assumption.
Upon associating a detection with a target, the detected bounding box is employed to update the target state. 
If no detection is associated with the target, its state is predicted straightforwardly using the
linear velocity model. DeepSORT adopts two types of distance metrics, namely IoU and appearance similarity. 
The association between the predicted Kalman states and new detections is resolved 
through the matching cascade, treating the association task as a sequence of sub-problems
rather than a global assignment problem. The core idea of it is to accord higher matching priority 
to more frequently seen objects.
Each association sub-problem is resolved using the Hungarian algorithm.
\\
\textbf{The State Space} \ The state of each target is modelled as $(x,y,\gamma,h,\dot{x},\dot{y},\dot{\gamma}, \dot{h})$, 
representing the
bounding box center position $(x, y)$, the aspect ratio $\gamma$, the height \emph{$h$}, and their respective velocities in image coordinates.
\\
\textbf{Deep Appearance Descriptor} \ DeepSORT utilizes a simple CNN pre-trained
on the person ReID dataset MARS~\cite{zheng2016mars} to extract an appearance descriptor 
$r_j$ for each detected bounding box. 
The distance between appearance descriptors could be used as the matching cost during the association procedure.
The primary challenge for ReID models lies in maintaining feature robustness against 
varying poses, low resolution, different backgrounds, 
and poor illumination and so on. To mitigate this issue, a
feature gallery $G_i$=$\{r^{(k)}_i\}_{k=1}^{N}$ is employed by DeepSORT to store the latest $N$=100
associated appearance descriptors for each track $i$. Then, the appearance similarity
between the $i$-th track and $j$-th detection is computed as
\begin{equation}
    d(i,j) = min{\{1-r_j^Tr_i^{(k)} \mid r_i^{(k)} \in G_i\}}.
    \label{eq:deepsort_reid}
\end{equation}

While the memory technology enhances the robustness and effectiveness of the ReID model, 
it concurrently consumes a considerable amount of memory and computing time, 
making it less conducive to real-time tracking. It is noteworthy that DeepSORT 
integrates motion information with appearance similarity through Mahalanobis 
distance, serving as a filter to exclude improbable associations.
\\
\textbf{Creation and Deletion of Track Identities} \ DeepSORT categorizes tracks into three classes: tentative, confirmed, and deleted.
As new objects enter the image, unique identities are created and labelled as tentative, 
reflecting the uncertainty of the detector. 
After the new tracks are associated with detections successfully for a sufficient number of times,
the tentative tracks are labelled as confirmed. Tracks are terminated immediately if they are not detected 
to prevent an uncontrolled increase in the number
of tracks and localization errors resulting from prolonged predictions without corrections 
from the detections. However, this causes DeepSORT to lose the ability to re-identify 
the reappeared object.

\section{MapTrack}

\begin{algorithm*}[ht]
    \caption{Overall Procedure of MapTrack}
    \label{alg:MapTrack}
    \renewcommand{\algorithmicrequire}{\textbf{Input:}}
    \renewcommand{\algorithmicensure}{\textbf{Output:}}
    \begin{algorithmic}[1]
        \REQUIRE Tentative tracks set $T_{tentative}$, normal tracks set $T_{normal}$, predicted tracks set $T_{predicted}$, disappeared tracks set $T_{disappeared}$, detections set $D$, matched tracks and detections set $M=\emptyset$  
        \ENSURE Updated tentative tracks set $T_{tentative}$, updated normal tracks set $T_{normal}$, updated predicted tracks set $T_{predicted}$, updated disappeared tracks set $T_{disappeared}$    
        
        \STATE $D$ $\gets$ \emph{Detections\_Filtering\_Stage\_1($D$)}
        \STATE The prediction stage of the Kalman filter
        \STATE $T_{tentative}$ $\gets$ \emph{Tentative}\_\emph{Tracks}\_\emph{Filtering}\_\emph{Stage\_1}($T_{tentative}$, $T_{normal}$, $T_{predicted}$)
        \STATE $M_{tentative}$, $UT_{tentative}$, $UD_{tentative}$ $\gets$ \emph{IoU}\_\emph{association}$(T_{tentative}, D)$, where $M_{tentative}$ represents matched tracks and detections, $UT_{tentative}$ represents unmatched tracks, $UD_{tentative}$ represents unmatched detections (the same below)
        \STATE Delete $UT_{tentative}$
        \STATE $M_{normal}$, $UT_{normal}$, $UD_{normal}$ $\gets$ \emph{IoU}\_\emph{ReID}\_\emph{association}($T_{normal}$, $D$). Note that \emph{IoU}\_\emph{ReID}\_\emph{association} requires matched associations are both verified by IoU and appearance metrics
        \STATE $M$, $UD$ $\gets$ \emph{Tentative\_Tracks\_Filtering\_Stage\_2}($M_{normal}$, $M_{tentative}$, $UD_{normal}$)
        \STATE Divide $UT_{normal}$ into crowded ones $UT_{normal\_crowded}$ and uncrowded ones $UT_{normal\_uncrowded}$ using the prediction map
        \STATE $M_{normal\_uncrowded}$, $UT_{normal\_uncrowded\_IoU}$, $UD$ $\gets$ \emph{IoU}\_\emph{association}($UT_{normal\_uncrowded}, UD$)
        \STATE $M_{normal\_crowded}$, $UT_{normal\_crowded\_ReID}$, $UD$ $\gets$ \emph{ReID}\_\emph{association}($UT_{normal\_crowded}, UD$)
        \STATE $M$, $UT_{normal\_crowded\_ReID}$ $\gets$ \emph{Detections}\_\emph{Filtering}\_\emph{Stage}\_\emph{2}($T_{predicted}$, $UT_{normal\_crowded}$, $M_{normal\_crowded}$, $M$, $M_{normal\_uncrowded}$)
        \STATE $UT_{normal}$ = $UT_{normal\_uncrowded\_IoU}$ + $UT_{normal\_crowded\_ReID}$
        \STATE $UD$ $\gets$ \emph{Detections\_Filtering\_Stage\_3}($UT_{normal}$, $T_{predicted}$, $UD$)
        \STATE $M_{predicted\_IoU}$, $UT_{predicted\_IoU}$, $UD_{predicted\_IoU}$ $\gets$ \emph{IoU\_association}($T_{predicted}$, $UD$)
        \STATE $M_{predicted\_ReID}$, $UT_{predicted\_ReID}$, $UD_{predicted\_ReID}$ $\gets$ \emph{ReID\_association}($T_{predicted}$, $UD$)
        \STATE $M$, $UD$, $UT_{predicted}$ $\gets$ \emph{Predicted}\_\emph{Tracks}\_\emph{Filtering}($M_{predicted\_ReID}$, $M_{predicted\_IoU}$, $M$)
        \STATE $M_{disappeared}$, $UT_{disappeared}$, $UD$ $\gets$ \emph{ReID\_association}($T_{disappeared}$, $UD$)
        \STATE $M$ = $M$ + $M_{disappeared}$
        \STATE $T_{tentative}$, $T_{normal}$, $T_{predicted}$, $T_{disappeared}$ $\gets$ \emph{Post\_processing}($M$, $UD$, $UT_{normal}$, $UT_{predicted}$, $UT_{disappeared}$)
    \end{algorithmic}
\end{algorithm*}

\noindent Differing from DeepSORT, which categorizes tracks into three classes, 
we classify them as tentative, normal, predicted, disappeared, and deleted. 
The tentative tracks resemble those in DeepSORT, except that we design a filtering algorithm 
for tentative tracks 
to eliminate falsely generated instances due to the false positives of the detector or crowded scenes.
The tentative tracks transition to normal tracks after being consecutively associated with new detections 
for a specific period. Unassociated normal tracks are labelled as predicted or 
disappeared depending on whether they persist in the probability map.
Predicted tracks propagate their states using the linear velocity model when not associated 
with new detections. Disappeared
tracks have a chance to be matched by the ReID model if they reappear. We prioritize normal and 
tentative tracks in data association, followed by predicted and disappeared tracks in order. 
The entire process of our MapTrack is detailed in algorithm \ref{alg:MapTrack}.

As illustrated in algorithm \ref{alg:MapTrack}, different distance metrics and association strategies 
are employed for tentative, normal, predicted, and disappeared tracks. 
Specifically, tentative tracks are associated through IoU metric. Regarding normal 
tracks, we initially identify matches verified by both IoU and appearance metrics with high association thresholds 
to minimize mis-associations. Subsequently, remaining unmatched normal tracks
are associated by either appearance or IoU metrics, 
according to whether they are situated within a crowd or not. 

For predicted tracks, all matches from IoU or appearance metrics are accepted, with a 
preference for the appearance metric in cases of conflict. This preference is driven 
by two factors\cite{observation_centric}.
Firstly, even minor positional shifts can result in substantial variations in estimated speed, 
leading to significant deviations in localization. Secondly, 
the rapid accumulation of estimation errors occurs in the consecutive absence of observations. 
We only apply appearance 
metric for disappeared tracks
due to the unavailability of location information.

While the matching cascade algorithm plays a non-trivial role in DeepSORT, 
we substitute it with a global assignment approach for three reasons. 
Firstly, we do not utilize the Mahalanobis distance
between predicted Kalman states and new detections~\cite{deepsort}. 
Secondly, it's not reasonable to give 
priority to more frequently seen objects in many scenes. 
Finally, our MapTrack is more robust to occluded
and crowded scenes, and additional prior constraints lead to more mis-associations instead.

\subsection{Covariance Adaptive Kalman Filter}
\noindent Different from DeepSORT, the state space of our MapTrack is modelled as $(x,y,w,h,\dot{x},\dot{y})$. 
Specifically, the aspect ratio $\gamma$
is directly substituted with the width of the bounding box $w$, and the velocities of the aspect ratio and the height are omitted. 
This omission is made because 
the width and height of the detected bounding box fluctuate randomly due to the 
inherent nature of the detector. Thus, the removal prevents the rapid and unbounded growth 
or decrease in the width and height caused by prolonged predictions
without corrections from detections.
\\
\textbf{CA Kalman Filter} \ In DeepSORT, the measurement covariance is constant. However, intuitively the measurement 
uncertainties escalate in the crowded and occluded scenes. Furthermore, the measurement covariance 
should be adjusted based on the degree of deformation observed in the detected bounding boxes.  
Therefore, we develop a formula to adaptively modulate the measurement covariance, denoted as the CA measurement covariance $\tilde{R}_i$ for each track $i$:
\begin{equation}
    \tilde{R}_i=f(d_i)R,
    \label{eq:measurement_covariance}
\end{equation}
where $f(x)$ is a piecewise function:
\begin{equation}
    {f(x)}=
        \begin{cases}
        coef1 &x \in (1.4,+\infty) \cup (0,0.6) \\
        coef2 &x \in (1.3,1.4) \cup (0.6,0.7) \\
        coef3 &x \in (1.2,1.3) \cup (0.7,0.8). \\
        \end{cases}
    \label{eq:piecewise_function}
\end{equation}
And $R$ is the preset constant measurement covariance and $d_i$ represents the degree of deformation in
the detected bounding boxes, which is calculated according to the ratio of the area of the detected bounding box to the area of the predicted bounding box:
\begin{equation}
    d_i= \frac{A_i^d}{A_i^p},
    \label{eq:degree_of_deformation}
\end{equation}
where $A_i^p$ is the area of the predicted box for track $i$, and $A_i^d$ is the area of the associated detected bounding box.
\\
\textbf{Momentum of Velocity} \ 
We use the momentum strategy to aid the measurement of velocity. The velocity of 
the $i$-th track in the $j$-th frame is measured by:
\begin{equation}
    v_i^j= \beta v_i^{j-1} + (1-\beta)\hat{v}_i^j,
    \label{eq:momentum_of_velocity}
\end{equation}
where $\beta$ denotes the momentum term, and $\hat{v}_i^j$ represents the computed velocity under the constant velocity assumption in the $j$-th frame:

\begin{equation}
    \hat{v}_i^j= \frac{p_i^j-p_i^{j-1}}{\Delta t},
    \label{eq:constant_velocity}
\end{equation}
where $p_i^j$ denotes the detected box center associated with the $i$-th track in the $j$-th frame,
while $p_i^{j-1}$ represents the updated box center of the $i$-th track in the preceding frame. The symbol $\Delta t$ 
signifies the time interval between consecutive frames.

\subsection{Probability Map and Prediction Map}
\noindent As previously mentioned, MapTrack excels in the handling of crowded scenes and occlusion, 
which owes a great deal to the probability map and prediction map.
\\
\textbf{Probability Map}
The probability map is a matrix where both the columns and rows are one-tenth of the width and height of the frame in image coordinates.
All values within the matrix are initialized to zero. 
For each new detected bounding box, the values at its corresponding positions in the probability map are incremented by one. 
Consequently, each value within the probability map signifies the cumulative count the object has passed by, and can also be interpreted as a probability after normalization. 
How can we determine whether the tracks exist within the probability map? 
If the tracks are on the verge of leaving the image or entering regions where the sum 
of probabilities falls below a specified threshold $thresh1$, they are considered to be outside the 
bounds of the probability map.
\\
\textbf{Feature Repository}
A feature repository is implemented to store tracks which are identified as disappeared by the probability map. 
Upon their reappearance, 
a ReID model is employed to re-identify them, as depicted at step 17 of algorithm \ref{alg:MapTrack}.
Evidently, this mechanism incurs significant memory consumption. Therefore, it is imperative to 
devise a deletion strategy tailored to specific application scenarios to prevent uncontrolled 
growth in the number of tracks.
\\
\textbf{Prediction Map}
The prediction map shares the same dimensions as the probability map and is initialized to zeros 
for each data association process. Following the prediction stage of the Kalman filter, 
all normal and predicted tracks are then mapped onto the prediction map, 
with their corresponding values incremented by one.
For each track, if the ratio of the sum of its corresponding values on the prediction map to 
its area exceeds the threshold value $thresh2$, the track is considered to be in a crowded state.

\subsection{Crowds and Occlusion Handling}
\noindent The fundamental principle guiding the management of crowded and occluded situations 
is to prioritize predictions over detections in such scenes. On the one hand, 
if normal tracks, already in the probability map, fail to associate with new detections, 
it is reasonable to assume that they are temporarily undetected possibly due to either occlusion or 
a detector failure.
Subsequently, they are labelled as predicted.
On the other hand, significant deformations of detected bounding boxes commonly occur in crowded 
scenes, such as when only the upper body or a portion of the body is detected. 
This situation can result in missing or false associations. If we assign the same 
level of trust to detections in these scenarios as we do in normal cases during 
the update stage of the Kalman filter, substantial deviations will undoubtedly 
occur. Therefore, in crowded scenes, greater trust should be placed in predictions 
than in detections, 
which is accomplished by the CA Kalman filter.

\subsection{Detection and Track Filtering Algorithm}
\begin{figure}[!t]
    \centering
     \includegraphics[width=0.8\linewidth]{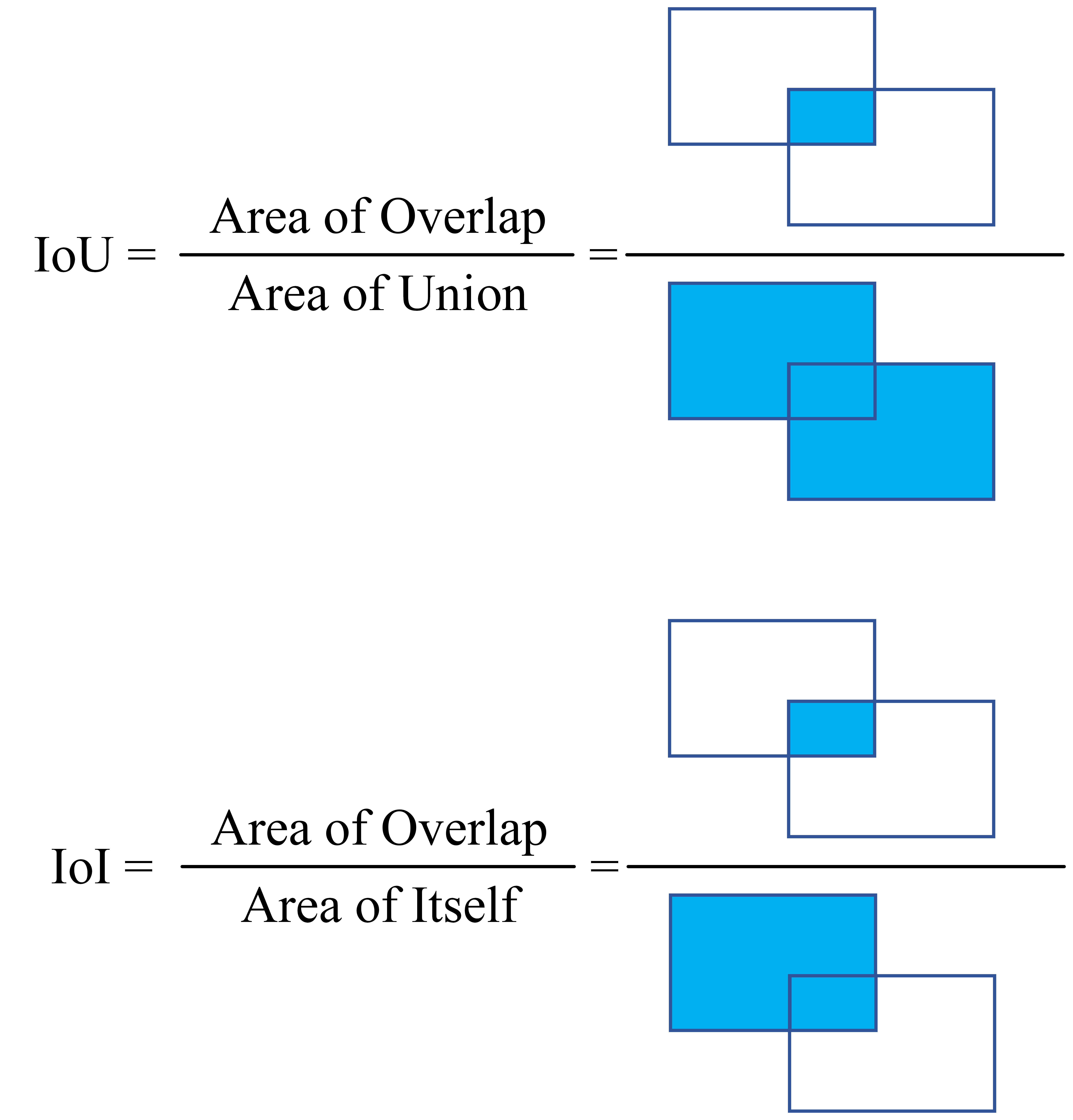}
  
     \caption{Illustration of how Intersection over Itself (IoI) is calculated and the comparison with Intersection over Union (IoU).}
     \label{fig:ioi}
  \end{figure}

\noindent The primary objective is to filter out falsely generated detections, tentative tracks, and  
predicted tracks resulting from the false positives of the detector and challenging scenes, 
such as crowded and occluded ones. 
To address this issue, a new metric is proposed, termed Intersection over Itself (IoI), 
representing the ratio of the intersection area to its own area (Figure \ref{fig:ioi}).

The detection filtering algorithm consists of three stages. 1. The first is step 1 of algorithm \ref{alg:MapTrack}, 
which filters out the detected bounding boxes with high IoI values. 
This process is undertaken to eliminate low-quality detections produced by the detector.
2. The second is step 11 of algorithm \ref{alg:MapTrack}. 
Following the preceding step, where ReID matching results are obtained, it is acknowledged 
that a resemblance in appearance may confuse the ReID model, resulting in ID switching. 
Thus, for each matched detected bounding box, 
we find its nearby normal and predicted tracks that have non-zero IoI values with it. 
Subsequently, appearance similarities are computed between 
the matched detected bounding box and its nearby track boxes. 
If the ReID values of two or more are sufficiently close, this match, 
along with the corresponding detection, is removed. 
3. Finally, the third stage corresponds to step 13 of algorithm \ref{alg:MapTrack}. 
For each remaining unmatched detected bounding box, IoI values are computed between it and 
the remaining 
unmatched normal and predicted tracks. 
If this detected bounding box has high IoI values with two or more tracks, 
we delete this detected bounding box. 
In essence, we remove detected boxes with more than one potential match, 
effectively preventing possible false matches.

There are two analogous stages in the tentative track filtering algorithm. 
The first stage corresponds to step 3 of algorithm \ref{alg:MapTrack}. 
For each tentative track, IoI values are computed between it and all normal and predicted tracks.
If this tentative track exhibits a high IoI value with two or more tracks, it is removed. 
This approach effectively prevents potential false transitions from tentative tracks to normal tracks.
The second stage corresponds to step 7 of algorithm \ref{alg:MapTrack}. 
At step 4 of algorithm \ref{alg:MapTrack}, matches between tentative tracks and the detection 
set \emph{$D$} are identified,
and at step 6, 
matches between normal tracks and the same detection set \emph{$D$} are determined.
Ideally, tentative and normal tracks should not be associated with the same detections.
However, in challenging scenes, this does occur. 
Therefore, we identify the detections that are associated with both tentative and normal 
tracks simultaneously, and eliminate the tentative tracks,
as they are more prone to be matched falsely compared to normal tracks.

Incorrectly predicted tracks could potentially lead to false associations. 
Hence, we have devised the predicted track filtering algorithm, 
corresponding to the step 16 of algorithm \ref{alg:MapTrack}.
At steps 14 and 15 of algorithm \ref{alg:MapTrack}, IoU and appearance 
similarity are utilized to respectively associate the predicted tracks 
with the same set of remaining unmatched 
detections.
In the predicted track filtering stage, matches are initially eliminated if the ratio of the distance 
between the predicted track and the associated detected bounding box to the width of the detected 
bounding box exceeds the threshold $thresh3$.
If one detected bounding box is associated with two different predicted tracks through 
IoU and appearance metrics respectively, greater trust is placed in the appearance metric than in IoU.
Subsequently, the predicted tracks are either deleted or labelled as disappeared under 
three circumstances. Firstly, if they are not encompassed within the probability map. 
Secondly, if they are not situated within the crowd. 
Thirdly, if they are far away from normal tracks, where IoI serves as the distance metric.

You might be wondering whether the detection and track filtering could lead to false deletions. 
Our response is that you need not worry for three reasons. Firstly, incorrectly deleted 
detected boxes and tentative tracks would reappear, and our prediction-based strategy 
would come into effect to prevent ID switching. Secondly, unmatched normal, predicted, 
or disappeared tracks resulting from mistakenly deleted detections would transition to 
predicted tracks or remain unchanged, rather than being deleted. Thirdly, incorrectly 
deleted predicted tracks have negligible impact on tracking performance.

\subsection{Post-processing}
\noindent In the post-processing step, we mainly do three things. 
Firstly, update the states of matched tracks and propagate the states
of the predicted tracks by the linear velocity model.
Secondly, initialize the unmatched detections as new tentative tracks. 
Finally, label the remaining unmatched normal tracks as predicted or disappeared 
based on their presence in the probability map.

\section{Experiments}

\begin{table*}[ht]
  \caption {Comparison with the state-of-the-art methods under the “public detector” protocol on the MOT17 test set. $\uparrow$ means higher is
  better, $\downarrow$ means lower is better. Bold and underlined numbers indicate the best and the
  second best performances.}
  \label{tab:mot17_public}
	\centering
	\begin{tabular}[c]{ccccccccc}
		\toprule
		{Tracker} & {Tracking Mode} & {HOTA$\uparrow$} & {MOTA$\uparrow$} & {IDF1$\uparrow$} & {AssA$\uparrow$} & {DetA$\uparrow$} & {IDSW$\downarrow$} & {Frag$\downarrow$} \\
		\midrule
		GMPHD-KCF~\cite{kutschbach2017sequential} & online & 30.3 & 39.6 & 36.6 & 26.4 & 35.3 & 5811 & 7414 \\
    	HISP-DAL~\cite{baisa2021robust} & online & 34.0 & 45.4 & 39.9 & 30.7 & \underline{38.1} & 8727 & \underline{7174} \\
    	MASS~\cite{karunasekera2019multiple} & online & 36.8 & \underline{46.9} & 46.0 & 35.4 & \textbf{38.5} & 4478 & 11994 \\
		GM-PHD-DAL~\cite{baisa2019online} & online & 31.4 & 44.4 & 36.2 & 26.6 & 37.5 & 11137 & 13900 \\
		GM-PHD-N1Tr~\cite{baisa2019development} & online & 30.3 & 42.1 & 33.9 & 25.9 & 36.1 & 10698 & 10864 \\
		OneShotDA~\cite{yoon2020oneshotda} & online & \underline{36.9} & \textbf{51.4} & \textbf{54.0} & \textbf{43.5} & 31.4 & \textbf{2118} & \textbf{3072} \\
		MapTrack (ours) & online & \textbf{37.6} & 40.6 & \underline{48.1} & \underline{40.0} & 35.7 & \underline{3329} & 13936 \\
		\bottomrule
	\end{tabular}
	
\end{table*}

\begin{table*}[ht]
	\centering
  \caption {Comparison with the state-of-the-art methods under the “public detector” protocol on the MOT20 test set. $\uparrow$ means higher is
  better, $\downarrow$ means lower is better. Bold and underlined numbers indicate the best and the
  second best performances.}
  \label{tab:mot20_public}
	\begin{tabular}[c]{ccccccccc}
		\toprule
		{Tracker} & {Tracking Mode} & {HOTA$\uparrow$} & {MOTA$\uparrow$} & {IDF1$\uparrow$} & {AssA$\uparrow$} & {DetA$\uparrow$} & {IDSW$\downarrow$} & {Frag$\downarrow$} \\
		\midrule
		OVBT~\cite{ban2016tracking} & online & 30.5 & 40.0 & \underline{37.8} & \underline{27.4} & 34.4 & \textbf{4210} & \underline{10026} \\
		CTv0~\cite{lohn2022clustertracker} & offline & \underline{33.0} & \textbf{45.1} & 35.6 & 26.3 & \textbf{42.3} & 6492 & \textbf{6351} \\
		MapTrack (ours) & online & \textbf{33.6} & \underline{40.7} & \textbf{42.4} & \textbf{32.5} & \underline{35.0} & \underline{4383} & 16621 \\
		\bottomrule
	\end{tabular}

\end{table*}

\begin{table*}[ht]
	\centering
  \caption {Comparison with the \textbf{reproduced} state-of-the-art methods on the MOT17 validation set. $\uparrow$ means higher is
  better, $\downarrow$ means lower is better. Bold and underlined numbers indicate the best and the
  second best performances.}
  \label{tab:mot17_reproduce}
	\begin{tabular}[c]{ccccccccccc}
		\toprule
		{Tracker} & {Tracking Mode} & {HOTA$\uparrow$} & {MOTA$\uparrow$} & {IDF1$\uparrow$} & {AssA$\uparrow$} & {DetA$\uparrow$} & {IDSW$\downarrow$} & {Frag$\downarrow$} \\
		\midrule
		ByteTrack~\cite{zhang2022bytetrack} & online & 46.0 & \underline{50.7} & 58.9 & 48.9 & 43.8 & \underline{109} & \textbf{284} \\
		BoT-SORT~\cite{aharon2022bot} & online & \underline{46.5} & \textbf{51.5} & \underline{60.0} & \underline{49.7} & \underline{44.1} & \textbf{96} & \underline{294} \\
		DeepSORT~\cite{deepsort}  & online & 44.4 & 49.2 & 56.1 & 45.0 & \textbf{44.6} & 273 & 749 \\
		MapTrack (ours) & online & \textbf{47.7} & 47.7 & \textbf{62.3} & \textbf{52.3} & 44.0 & 178 & 940 \\
		\bottomrule
	\end{tabular}
	
\end{table*}

\begin{table*}[ht]
	\centering
  \caption {Comparison with the \textbf{reproduced} state-of-the-art methods on the MOT20 validation set. $\uparrow$ means higher is
  better, $\downarrow$ means lower is better. Bold and underlined numbers indicate the best and the
  second best performances.}
  \label{tab:mot20_reproduce}
	\begin{tabular}[c]{ccccccccccc}
		\toprule
		{Tracker} & {Tracking Mode} & {HOTA$\uparrow$} & {MOTA$\uparrow$} & {IDF1$\uparrow$} & {AssA$\uparrow$} & {DetA$\uparrow$} & {IDSW$\downarrow$} & {Frag$\downarrow$} \\
		\midrule
		ByteTrack~\cite{zhang2022bytetrack} & online & 30.4 & 44.4 & 43.6 & \underline{28.7} & 32.8 & \underline{2141} & \underline{6786} \\
		BoT-SORT~\cite{aharon2022bot} & online & \underline{30.9} & 45.3 & \underline{44.3} & \textbf{29.3} & 33.2 & \textbf{2080} & \textbf{6627} \\
		DeepSORT~\cite{deepsort}  & online & 25.0 & \textbf{48.4} & 32.6 & 17.8 & \textbf{35.8} & 5476 & 14880 \\
		MapTrack (ours) & online & \textbf{31.1} & \underline{46.8} & \textbf{45.3} & 28.6 & \underline{34.3} & 3971 & 20212 \\
		\bottomrule
	\end{tabular}
	
\end{table*}

\subsection{Datasets and Metrics}
\noindent We evaluate our MapTrack under the “public
detection” protocol on the challenging pedestrian
tracking MOT17~\cite{milan2016mot16} and MOT20~\cite{dendorfer2020mot20}
datasets. 
MOT17 comprises 7
sequences with 5316 frames for training and 7 sequences with
5919 frames for testing, featuring frequent occlusions and crowded scenes. 
MOT20 focuses on extremely crowded scenes,
comprising 4 sequences with 8931 frames for training and 4 sequences with 4479 frames for testing.
These sequences exhibit variations in angle of view, object size, camera motion, and frame rate.
We select randomly half of the training sequences of both MOT17 and MOT20 for training and the second half for validation.

The Higher Order Tracking Accuracy (HOTA)~\cite{luiten2021hota} is chosen as the primary 
metric due to its balanced assessment of object detection and association accuracy, 
defined as the geometric 
mean of detection accuracy (DetA) and association accuracy (AssA). 
Complementary to this, we emphasize additional metrics to 
comprehensively evaluate various aspects of tracking performance. 
The Multi Object Tracking Accuracy (MOTA)~\cite{bernardin2008evaluating} serves as a measure of overall tracking performance,
evaluating errors from three sources:
false negatives (FN), false positives (FP) and identity switches (IDSW). MOTA is highly related to detection performance
because in general the number of FPs and FNs exceeds IDSW significantly.
The ID F1 Score (IDF1)~\cite{ristani2016performance} places a specific focus on association performance 
by assessing the tracker's ability to sustain identities over time.

\subsection{Implementation Details}
\noindent We implemented MapTrack in Python and conducted all experiments and training on
a single NVIDIA GeForce RTX 3060 Ti GPU.
To ensure a fair and meaningful comparison with other tracking methods, 
as strongly advocated by the MOT Benchmark officially, we present all results on the public
detections provided by MOT17 and MOT20. In the update stage of the Kalman filter for the tentative and normal
tracks, we set $coef1$=15, $coef2$=9, and $coef3$=6. While for the predicted tracks, we set
$coef1$=9, $coef2$=6, and $coef3$=3. And we set $\beta$=0.9, $thresh1$=0.05, $thresh2$=1.25, $thresh3$=3 respectively.
Considering that some state-of-the-art methods did not provide results on the public detections,
and for a more thorough and comprehensive comparison, ByteTrack~\cite{zhang2022bytetrack}, 
DeepSORT~\cite{deepsort}, and BoT-SORT~\cite{aharon2022bot} were reproduced 
using the same detector and ReID model as MapTrack. 
The rationale behind selecting these methods for comparison lies in that they are 
also DeepSORT-like or SORT-like frameworks.
For object detection, we
adopt YOLOv8m as the detector, pre-trained on the COCO dataset~\cite{lin2015microsoft}, 
for its balance
between accuracy and speed. Additionally, fine-tuning was conducted using the WiderPerson 
dataset~\cite{zhang2019widerperson}. 
The input image
size is 640 $\times$ 640. The object confidence threshold for detection is consistently set 
to 0.25 across all datasets, except for ByteTrack, where the high and low object confidence 
thresholds are specified as 0.5 and 0.1, respectively. 
For the ReID model, we employ the lightweight yet efficient OSNet~\cite{zhou2019omni} model, 
which is pre-trained on the MSMT17 dataset~\cite{wei2018person}. Our MapTrack is 
then compared to these reproduced approaches on the validation sets of MOT17 and MOT20.

\subsection{Comparison with the State-of-the-Art Methods}
\noindent \textbf{Public Detections} \ We conduct a comparative analysis of MapTrack against other 
state-of-the-art methods, utilizing the public detections on the test datasets of MOT17 and MOT20, 
as presented in Table \ref{tab:mot17_public} and Table \ref{tab:mot20_public}, respectively.
MapTrack demonstrates superior performance, surpassing other methods in HOTA 
and securing the second position for IDF1, AssA, and IDSW on MOT17.
On MOT20, it attains the top rank in HOTA, IDF1, and AssA, while achieving second place for MOTA, DetA, and IDSW.

MapTrack is designed with a primary focus on enhancing robustness in the presence of the crowd 
and occlusion, excelling at identity preservation. 
This emphasis contributes to our competitive
results in metrics such as HOTA, IDF1, AssA, and IDSW that reflect the association performance.
\\
\textbf{Private Detections} \
We conduct a comparative analysis of MapTrack against several prominent state-of-the-art methods, 
utilizing YOLOv8 as the detector and OSNet as the ReID model
on the validation set of MOT17 and MOT20, as presented
in Table \ref{tab:mot17_reproduce} and Table \ref{tab:mot20_reproduce} respectively.

DeepSORT serves as our baseline approach. On MOT17, it is evident that MapTrack improves the HOTA metric
of DeepSORT from 44.4 to 47.7, IDF1 from 56.1 to 62.3, AssA from 45.0 to 52.3, and
decreases IDSW from 273 to 178. On MOT20, MapTrack similarly enhances the HOTA metric
of DeepSORT from 25.0 to 31.1, IDF1 from 32.6 to 45.3, AssA from 17.8 to 28.6, and
decreases IDSW from 5476 to 3971. 
The improvements on the metrics that mainly measure the association performance, 
and the fact that the improvements on MOT20 are larger than those on MOT17 
prove and highlight the enhanced robustness of MapTrack over crowds and occlusion.
It is worth noting that
our reproduced version of DeepSORT also performs
well on the benchmark, demonstrating the effectiveness
of the DeepSORT-like tracking paradigm.

Our MapTrack surpasses BoT-SORT and ByteTrack in terms of HOTA, IDF1, and AssA on MOT17, 
as well as HOTA and IDF1 on MOT20.
These results further highlight MapTrack's robustness in handling crowded and occluded scenes.
You may be wondering that why MapTrack does not exhibit strong performance in terms of
IDSW and Frag, despite its improved association performance.  
This can be attributed to the utilization of the prediction-based strategy. 
In crowded and occluded scenarios, MapTrack places greater reliance on predictions over detections.
As a consequence, undetected normal tracks would transition to
the predicted tracks, leading to trajectory interruptions and the increase in the value of Frag.
Moreover, the trajectory interruptions may be misinterpreted as instances of ID switching.
That is to say, the somewhat diminished performance in terms of IDSW and Frag can be 
regarded as a price of achieving superior tracking performance.
It is important to note that identity switches and trajectory interruptions 
do not really occur, as predicted tracks would transition to normal 
tracks following association with new detections, 
thereby maintaining identities consistently highly possibly.

Our MapTrack ranks fourth in MOTA on MOT17. The primary factor contributing to 
this ranking, apart from the relatively sub-optimal performance in IDSW, is the increased  
incidence of false negatives. Unlike ByteTrack and BoT-SORT, which leverage low-confidence 
detections to achieve elevated MOTA scores, we refrain from employing adaptive detection thresholds. 
This decision is driven by the concern that
the tracking performance would be impaired
by a large number of false positives with a
lower confidence threshold.
To address this issue, specialized strategies, such as those employed in ByteTrack, are necessary. 
However, the primary objective of MapTrack is to demonstrate our innovative approaches.
And of course MapTrack can easily be generalized by equipping with other state-of-the-art techniques.

\section{Conclusion}
\noindent We first revisit the widely adopted and classic TBD paradigm, DeepSORT, 
and recognize its intrinsic limitations for 
ID switching in crowded and occluded scenes. In response to these challenges, 
an effective online and real-time MOT approach, named MapTrack, is introduced.
Maptrack introduces the probability map, the prediction map, and the covariance adaptive Kalman filter
to address the missing and false association problems in crowded and occluded scenes. 
The main idea is to place greater trust in predictions than detections when objects 
are undetected due to occlusions or when detected bounding boxes are distorted in crowded scenarios. 
The state-of-the-art performance on the MOT17 and MOT20 benchmark datasets 
has substantiated the advantages of our MapTrack. 
We hope our MapTrack can be deployed widely and contribute valuable insights to the community.

\bibliographystyle{IEEEtran}
\bibliography{IEEEabrv, ref}

\end{document}